# Learning Bayesian Networks: The Combination of Knowledge and Statistical Data

David Heckerman        Dan Geiger*        David M. Chickering

Microsoft Research, Bldg 9S
Redmond, WA 98052-6399
heckerma@microsoft.com, dang@cs.technion.ac.il, dmax@cs.ucla.edu

## Abstract

We describe scoring metrics for learning Bayesian networks from a combination of user knowledge and statistical data. We identify two important properties of metrics, which we call *event equivalence* and *parameter modularity*. These properties have been mostly ignored, but when combined, greatly simplify the encoding of a user's prior knowledge. In particular, a user can express his knowledge—for the most part—as a single *prior Bayesian network* for the domain.

## 1 Introduction

The fields of Artificial Intelligence and Statistics share a common goal of modeling real-world phenomena. Whereas AI researchers have emphasized a knowledge-based approach to achieving this goal, statisticians have traditionally emphasized a data-based approach. In this paper, we present a unification of the two approaches. In particular, we develop algorithms based on Bayesian principles that take as input (1) a user's prior knowledge expressed—for the most part—as a *prior Bayesian network* and (2) statistical data, and returns one or more improved Bayesian networks.

Several researchers have examined methods for learning Bayesian networks from data, including Cooper and Herskovits (1991,1992), Buntine (1991), and Spiegelhalter et al. (1993) (herein referred to as CH, Buntine, and SDLC, respectively). These methods all have the same basic components: a scoring metric and a search procedure. The metric computes a score that is proportional to the posterior probability of a network structure, given data and a user's prior knowledge. The search procedure generates networks for evaluation by the scoring metric. These methods use the two components to identify a network or set of networks with high posterior probabilities, and these networks are then used to predict future events.

In this paper, we concentrate on scoring metrics. Although we restrict ourselves to domains containing only discrete variables, as we show in Geiger and Heckerman (1994), our metrics can be extended to domains containing continuous variables. A major contribution of this paper is that we develop our metrics from a set of consistent properties and assumptions. Two of these, called parameter modularity and event equivalence, have been ignored for the most part, and their combined ramifications have not been explored. The assumption of *parameter modularity,* which has been made implicitly by CH, Buntine, and SDLC, addresses the relationship among prior distributions of parameters for different Bayesian-network structures. The property of *event equivalence* says that two Bayesian-network structures that represent the same set of independence assertions should correspond to the same event and therefore receive the same score. We provide justifications for these assumptions, and show that when combined with assumptions about learning Bayesian networks made previously, we obtain a straightforward method for combining user knowledge and statistical data that makes use of a prior network. Our approach is to be contrasted with those of CH and Buntine who do not make use of a prior network, and to those of CH and SDLC who do not satisfy the property of event equivalence.

Our identification of the principle of event equivalence arises from a subtle distinction between two types of Bayesian networks. The first type, called *belief networks*, represents only assertions of conditional independence. The second type, called *causal networks*, represents assertions of cause and effect as well as assertions of independence. In this paper, we argue that metrics for belief networks should satisfy event equivalence, whereas metrics for causal networks need not.

---

*Author's primary affiliation: Computer Science Department, Technion, Haifa 32000, Israel.

Our score-equivalent metric for belief networks is similar to metrics described by York (1992), Dawid and Lauritzen (1993) and Madigan and Raferty (1994), except that our metric scores directed networks, whereas their metrics score undirected networks. In this paper, we concentrate on directed models rather than on undirected models, because we believe that users find the former easier to build and interpret.

## 2 Belief Networks and Notation

Consider a domain $U$ of $n$ discrete variables $x_1, \ldots, x_n$. We use lower-case letters to refer to variables and upper-case letters to refer to sets of variables. We write $x_i = k$ when we observe that variable $x_i$ is in state $k$. We use $p(x = i|y = j, \xi)$ to denote the probability of a person with background knowledge $\xi$ for the observation $x = i$, given the observation $y = j$. When we observe the state for every variable in set $X$, we call this set of observations an *instance* of $X$. We use $p(X|Y, \xi)$ to denote the set of probabilities for all possible observations of $X$, given all possible observations of $Y$. The *joint space* of $U$ is the set of all instances of $U$. The *joint probability distribution* over $U$ is the probability distribution over the joint space of $U$.

A belief network—the first of the two types of Bayesian networks that we consider—represents a joint probability distribution over $U$ by encoding assertions of conditional independence as well as a collection of probability distributions. From the chain rule of probability, we know

$$p(x_1, \ldots, x_n|\xi) = \prod_{i=1}^{n} p(x_i|x_1, \ldots, x_{i-1}, \xi) \quad (1)$$

For each variable $x_i$, let $\Pi_i \subseteq \{x_1, \ldots, x_{i-1}\}$ be a set of variables that renders $x_i$ and $\{x_1, \ldots, x_{i-1}\}$ conditionally independent. That is,

$$p(x_i|x_1, \ldots, x_{i-1}, \xi) = p(x_i|\Pi_i, \xi) \quad (2)$$

A belief network is a pair $(B_S, B_P)$, where $B_S$ is a belief-network structure that encodes the assertions of conditional independence in Equation 2, and $B_P$ is a set of probability distributions corresponding to that structure. In particular, $B_S$ is a directed acyclic graph such that (1) each variable in $U$ corresponds to a node in $B_S$, and (2) the parents of the node corresponding to $x_i$ are the nodes corresponding to the variables in $\Pi_i$. (In the remainder of this paper, we use $x_i$ to refer to both the variable and its corresponding node in a graph.) Associated with node $x_i$ in $B_S$ are the probability distributions $p(x_i|\Pi_i, \xi)$. $B_P$ is the union of these distributions. Combining Equations 1 and 2, we see that any belief network for $U$ uniquely determines a joint probability distribution for $U$. That is,

$$p(x_1, \ldots, x_n|\xi) = \prod_{i=1}^{n} p(x_i|\Pi_i, \xi) \quad (3)$$

A *minimal belief network* is a belief network where Equation 2 is violated if any arc is removed. Thus, a minimal belief network represents both assertions of independence and assertions of dependence.

## 3 Metrics for Belief Networks: Previous Work

In this section, we summarize previous work, presented—for example—in CH, Buntine, and SDLC on the computation of a score for a belief-network structure $B_S$, given a set of cases $D = \{C_1, \ldots, C_m\}$. Each *case* $C_i$ is the observation of one or more variables in $U$. We sometimes refer to $D$ as a *database*.

A Bayesian measure of the goodness of a belief-network structure is its posterior probability given a database:

$$p(B_S|D, \xi) = c \; p(B_S|\xi) \; p(D|B_S, \xi)$$

where $c = 1/p(D|\xi) = 1/\sum_{B_S} p(B_S|\xi) \; p(D|B_S, \xi)$ is a normalization constant. For even small domains, however, there are too many network structures to sum over in order to determine the constant. Therefore researchers have used $p(B_S|\xi) \; p(D|B_S, \xi) = p(D, B_S|\xi)$ as a network-structure score. We note that this metric treats all variables as being equally important, but can be generalized [Spiegelhalter et al., 1993].

To compute $p(D, B_S|\xi)$ in closed form, researchers typically have made five assumptions, which we explicate here.

**Assumption 1** *The database $D$ is a multinomial sample from some belief network $(B_S, B_P)$.*

There are several assumptions implicit in Assumption 1. One is that all variables in $U$ are discrete. We modify this assumption in another paper in this proceedings [Geiger and Heckerman, 1994]. Another assumption is that the user may be uncertain as to which belief-network structure is generating the data. This uncertainty is encoded in the *prior probabilities for network structure* $p(B_S|\xi)$. Also implicit is that, given the data comes from a particular network structure, the user may be uncertain about the probabilities for that structure. These probabilities actually should be thought of as being long-run fractions that we would see in a very large database, and are called *parameters* in the statistical literature. Finally, we note that Assumption 1 implies that the processes generating the data do not change in time.

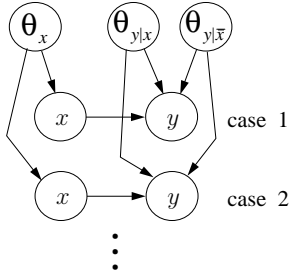

Figure 1: Illustration of Assumptions 1 and 2 for the network structure $x \to y$, where $x$ and $y$ are binary.

Assumption 1 can be represented in a belief network. Figure 1 illustrates the assumption for the network structure $x \to y$ where $x$ and $y$ are binary variables. (We shall use this two-variable domain to illustrate many of the points in this paper.) The parameter $\theta_x$ represents the long-run fraction of cases where $x$ is observed to be true. Given $\theta_x$, the observations of $x$ in each case are independent. The parameters $\theta_{y|x}$ and $\theta_{y|\bar{x}}$ represent the long-run fraction of cases where $y$ is observed to be true, in those cases where $x$ is observed to be true and false, respectively. If these two parameters are known, then the observations of $y$ in any two cases are independent, provided $x$ is also observed for at least one of those cases.

In general, given a belief-network structure $B_S$ for $U = \{x_1, \ldots, x_n\}$, we use $r_i$ to denote the number of states of variable $x_i$, and $q_i = \prod_{x_l \in \Pi_i} r_l$ to denote the number of instances of $\Pi_i$. We use the integer $j$ to index these instances. That is, we write $\Pi_i = j$ to denote the observation of the $j$th instance of the parents of $x_i$. We use $\theta_{ijk}$ to denote the long-run fraction of cases where $x_i = k$, in those cases where $\Pi_i = j$. We use $\Theta_{ij}$ to denote the union of $\theta_{ijk}$ over $k$, and $\Theta_{B_S}$ to denote the union of $\Theta_{ij}$ for all instances $j$ of all variables $x_i$. Thus, the set $\Theta_{B_S}$ corresponds to the parameter set $B_P$ for belief-network structure $B_S$, as defined in Section 2. Here, however, these parameters are long-run fractions whose values are uncertain. Also, we use $\rho(\cdot|\xi)$ to denote the probability density for a continuous variable or set of variables. For example, $\rho(\Theta_{ij}|B_S, \xi)$ denotes the probability density for the set of parameters $\Theta_{ij}$, given $B_S$ and $\xi$.

The next assumption, which we call *parameter independence*, says that the parameters associated with a given belief-network structure are independent, except for the obvious dependence among the parameters for a given variable (which must sum to one).

**Assumption 2 (Parameter Independence)** *For all belief-network structures $B_S$,*

$$\rho(\Theta_{B_S}|B_S, \xi) = \prod_i \prod_j \rho(\Theta_{ij}|B_S, \xi)$$

This assumption is illustrated in Figure 1 for the network structure $x \to y$.

If all variables in a case are observed, we say that the case is *complete*. If all cases in a database are complete, we say that the database is *complete*.

**Assumption 3** *All databases are complete.*

We note that Spiegelhalter et al. (1993) provide an excellent survey of approximations that circumvent this assumption.

A general metric now follows. Applying the chain rule, we obtain

$$p(D|B_S, \xi) = \prod_{l=1}^{m} p(C_l|C_1, \ldots, C_{l-1}, B_S, \xi) \quad (4)$$

where $C_i$ is the $i$th case in the database. Given Assumption 3, it follows that parameters remain independent when cases are observed. This conclusion is easily seen in the simple example of Figure 1.[1] Therefore, conditioning on the parameters of the belief-network structure $B_S$, we have

$$p(C_l|C_1, \ldots, C_{l-1}, B_S, \xi) = \quad (5)$$
$$\int_{\Theta_{B_S}} p(C_l|\Theta_{B_S}, B_S, \xi) \prod_i \prod_j \rho(\Theta_{ij}|C_1, \ldots, C_{l-1}, B_S, \xi)$$

Also, because each case in $D$ is complete, we have

$$p(C_l|\Theta_{B_S}, B_S, \xi) = \prod_i \prod_j \prod_k \theta_{ijk}^{\alpha_{lijk}} \quad (6)$$

where $\alpha_{lijk}$ is 1 if and only if $x_i = k$ and $\Pi_i = j$ in case $C_l$, and 0 otherwise. Plugging Equation 6 into Equation 5 and the result into Equation 4 yields

$$p(D, B_S|\xi) = p(B_S|\xi) \quad (7)$$
$$\cdot \prod_i \prod_j \prod_k \prod_l <\theta_{ijk}|C_1, \ldots, C_{l-1}, B_S, \xi>^{\alpha_{lijk}}$$

where $<\theta_{ijk}|\xi>$ denotes the expectation of $\theta_{ijk}$ with respect to $\rho(\Theta_{ij}|\xi)$.

One difficulty in applying Equation 7 is that, a user must provide prior distributions for every parameter set $\Theta_{ij}$ associated with every structure $B_S$. To reduce the number of prior distributions, we make the following assumption.

---
[1] In general, if a variable is observed in a belief network, we may delete all arcs emanating from it and retain a valid belief network.

**Assumption 4 (Parameter Modularity)**
*If $x_i$ has the same parents in any two belief-network structures $B_{S1}$ and $B_{S2}$, then for $j = 1, \ldots, q_i$,*

$$\rho(\Theta_{ij}|B_{S1}, \xi) = \rho(\Theta_{ij}|B_{S2}, \xi)$$

We call this property *parameter modularity,* because it says that the densities for parameters $\Theta_{ij}$ depend only on the structure of the belief network that is local to variable $x_i$—namely, $\Theta_{ij}$ only depends on the parents of $x_i$. For example, in our two-variable domain, let $B_{S1}$ be the network with an arc pointing from $x$ to $y$, and $B_{S2}$ be the network with no arc between $x$ and $y$. Then $\rho(\theta_x|B_{S1}, \xi) = \rho(\theta_x|B_{S2}, \xi)$ because $x$ has the same parents (namely, none) in both belief networks.

We note that CH, Buntine, and SDLC implicitly make the assumption of parameter modularity (Cooper and Herskovits, 1992, Equation A6, p. 340; Buntine, 1991, p. 55; Spiegelhalter et al., 1993, pp. 243-244). Also, in the context of causal networks, the assumption has a compelling justification (see Section 7). To our knowledge, however, we are the first researchers to make this assumption explicit. As we see in the following section, this assumption has important ramifications.

Given Assumption 3, parameter modularity holds even after cases have been observed. Consequently, we can drop the conditioning event $B_S$ in Equation 7, to yield

$$p(D, B_S|\xi) = p(B_S|\xi) \qquad (8)$$
$$\cdot \prod_i \prod_j \prod_k \prod_l <\theta_{ijk}|C_1, \ldots, C_{l-1}, \xi>^{\alpha_{lijk}}$$

In Heckerman et al. (1994), we provide greater detail about this general metric. Here, we concentrate on a special case where each parameter set $\Theta_{ij}$ has a Dirichlet distribution.

An important concept to be used in much of the remaining presentation is that of a complete belief network. A *complete belief-network* is one with no missing edges—that is, one that represents no assertions of conditional independence.

**Assumption 5** *For every complete belief-network structure $B_{S_C}$, and for all $\Theta_{ij} \subseteq \Theta_{B_{S_C}}$, $\rho(\Theta_{ij}|B_{S_C}, \xi)$ has a Dirichlet distribution. Namely, there exists exponents $N'_{ijk} > 0$, such that*

$$\rho(\Theta_{ij}|B_{S_C}, \xi) \propto \prod_k \theta_{ijk}^{N'_{ijk}-1}$$

From this assumption and our assumption of parameter modularity, it follows that for *every* belief-network structure $B_S$, and for all $\Theta_{ij} \subseteq \Theta_{B_S}$, $\rho(\Theta_{ij}|B_S, \xi)$ has a Dirichlet distribution.[2] When every such parameter set of $B_S$ has this distribution, we simply say that $\rho(\Theta_{B_S}|B_S, \xi)$ is Dirichlet.

Combining our previous assumptions with this consequence of Assumption 5, we obtain

$$\rho(\Theta_{ij}|D, B_S, \xi) \propto \prod_k \theta_{ijk}^{N'_{ijk}+N_{ijk}-1}$$

where $N_{ijk}$ is the number of cases in $D$ where $x_i = k$ and $\Pi_i = j$. Thus, if the prior distribution for $\Theta_{ij}$ has a Dirichlet distribution, then so does the posterior distribution for $\Theta_{ij}$. We say that the Dirichlet distribution is closed under multinomial sampling, or that the Dirichlet distribution is a *conjugate family* of distributions for multinomial sampling. Given this family,

$$<\theta_{ijk}|D, \xi> = \frac{N'_{ijk} + N_{ijk}}{N'_{ij} + N_{ij}} \qquad (9)$$

where $N_{ij} = \sum_{k=1}^{r_i} N_{ijk}$, and $N'_{ij} = \sum_{k=1}^{r_i} N'_{ijk}$. Substituting Equation 9 into each term of Equation 8, and performing the sum over $l$, we obtain

$$p(D, B_S^e|\xi) = p(B_S^e|\xi) \cdot \prod_{i=1}^{n} \prod_{j=1}^{q_i} \frac{\Gamma(N'_{ij})}{\Gamma(N'_{ij} + N_{ij})}$$
$$\cdot \prod_{k=1}^{r_i} \frac{\Gamma(N'_{ijk} + N_{ijk})}{\Gamma(N'_{ijk})} \qquad (10)$$

where $\Gamma$ is the *Gamma* function, which satisfies $\Gamma(x+1) = x\Gamma(x)$. We shall refer to Equation 10 as the BD metric (*B*ayesian metric with *D*irichlet priors), although we emphasize that this metric is not new.

Even with the inclusion of the assumption of parameter modularity, the application of this metric is difficult, because it requires that a user specify the Dirichlet exponents $N'_{ijk}$ for every complete belief network structure. In the following section, we introduce a property of belief-network metrics called *event equivalence*. In the subsequent section, we show how this property leads to a dramatic simplification of the assessment of these Dirichlet exponents.

## 4 Event Equivalence and Score Equivalence

In the previous section, we used $B_S$ as an argument of probabilities and probability densities. However, $B_S$ is a belief-network structure, not an event. Thus, we should have used $B_S^e$ in these situations, where $B_S^e$ is the event that corresponds to structure $B_S$ (the superscript "$e$" stands for *event*). In this section, we provide

---
[2]CH, Buntine, and SDLC express Assumption 5 in this form.

a definition of $B_S^e$ and explicate an important property of this definition.

A simple definition of $B_S^e$ is implicit in Assumption 1. In particular, this assumption says that (1) the database is a multinomial sample from the joint space of $U$, and (2) $B_S^e$ holds true iff the multinomial parameters for $U$ satisfy the independence assertions of $B_S$. For example, in our two-variable domain, Condition 1 corresponds to the assertion that a given database is a multinomial sample from the joint space $\{xy, x\bar{y}, \bar{x}y, \bar{x}\bar{y}\}$. Given $B_S$ is the network structure with no arc between $x$ and $y$, Condition 2 says that the event $B_S^e$ corresponds to the assertion $\theta_{xy} + \theta_{\bar{x}y} = \theta_{xy}/(\theta_{xy} + \theta_{x\bar{y}})$—that is, $\theta_y = \theta_{y|x}$.

This definition has the following desirable property. When two belief-network structures represent the same assertions of conditional independence, we say that they are *isomorphic*. For example, in the three variable domain $\{x, y, z\}$, the network structures $x \to y \to z$ and $x \leftarrow y \to z$ represent the same assertion: $x$ and $z$ are independent given $y$. Given the definition of $B_S^e$, it follows that events $B_{S1}^e$ and $B_{S2}^e$ are equivalent if and only if the structures $B_{S1}$ and $B_{S2}$ are isomorphic. That is, the relation of isomorphism induces an equivalence class on the set of events $B_S^e$. We call this property *event equivalence*.

There is a problem with the definition, however. In particular, events corresponding to non-isomorphic network structures are not mutually exclusive. For example, the event corresponding to a complete belief-network structure always holds true, and therefore implies the event corresponding to any other structure. In this case, and in general, the scores $p(D, B_S^e)$ associated with these network structures are useless for comparison.

A seemingly reasonable repair would be to say that $B_S^e$ holds true iff the multinomial parameters for $U$ satisfy the independence and *dependence* assertions of $B_S$, where $B_S$ is now interpreted as a *minimal* belief-network structure. Under this revised definition, each event is a set of equalities (as before), and also a set of inequalities. For example, given $B_{S1}$ is the belief-network structure $x \to y$ in our two-variable domain, then $B_{S1}^e$ is the event $\theta_{y|x} \neq \theta_y$; and given $B_{S2}$ is the belief-network structure with no arc, then $B_{S2}^e$ is the event $\theta_{y|x} = \theta_y$. These two events are mutually exclusive. Furthermore, the events corresponding to $x \to y$ and $y \to x$ are equal.

This repair is still not sufficient for larger domains, however. First, the property of score equivalence may be violated. For example, in the three-variable domain $\{x, y, z\}$, the events corresponding to complete belief-network structures for different orderings are not equal. We may recover this property by including in the event corresponding to a set of isomorphic network structures $E$ the union of inequalities associated with each such structure in $E$. Second, events corresponding to some non-isomorphic structures are not mutually exclusive. For example, the events corresponding to the structures $x \to y \leftarrow z$ and $x \to y \to z$ both include the situation where $\theta_{z|x} = \theta_z$. In general, however, such overlaps will be of measure zero with respect to the events that create the overlap. Thus, given a set of overlapping events, we may exclude the intersection from all but one of the events without affecting our mathematical results or the intuitive understanding of events by the user.

This revised definition of the event $B_S^e$ guarantees that the set of events corresponding to the set of all possible network structures for a given domain is mutually exclusive. Furthermore, the definition retains the property of event equivalence.

**Proposition 1 (Event Equivalence)**
*Belief-network structures $B_{S1}$ and $B_{S2}$ are isomorphic if and only if $B_{S1}^e = B_{S2}^e$.*

Because the score for network structure $B_S$ is $p(D, B_S^e|\xi)$, an immediate consequence of the property of event equivalence is score equivalence[3]:

**Proposition 2 (Score Equivalence)** *The scores of two isomorphic belief-network structures must be equal.*

We note that, given the property of event equivalence, we technically should score each belief-network-structure equivalence class, rather than each belief-network structure. Nonetheless, users find it intuitive to work with (i.e., construct and interpret) belief networks. Consequently, we continue our presentation in terms of belief networks, keeping Proposition 2 in mind.

It is easy to show that the BD metric given by Equation 10 does not exhibit the property of score equivalence for most choices of the Dirichlet exponents $N'_{ijk}$. Thus, the property of event equivalence must induce constraints on the parameters $N'_{ijk}$.

---
[3]In making the assumptions of parameter independence and parameter modularity, we have—in effect—specified the prior densities for the multinomial parameters in terms of the structure of a belief network. Consequently, there is the possibility that this specification violates the property of score equivalence. In Heckerman et al. (1994), however, we show that our assumptions and score equivalence are consistent.

# 5 The Prior Belief Network

In this section, we show how the property of event equivalence and Assumptions 1 through 5 lead to constraints on the exponents $N'_{ijk}$. We see that the constraints are so strong, that all exponents may be constructed from (1) a belief network reflecting the user's current knowledge about the next case, and (2) and equivalent sample size for the domain as a whole.

To begin, let us see how the property of event equivalence and Assumptions 1 through 4 constrain the prior densities $\rho(\Theta_{B_S}|B_S^e, \xi)$. That is, for the moment, let us ignore the assumption that densities are Dirichlet.

First, consider only complete belief-network structures. From the property of event equivalence, we know that the event associated with any complete belief-network structure for a given domain $U$ is the same; and we use $B_{S_C}^e$ to denote this event. So, suppose that we know the density of the multinomial parameters for the joint space of $U$ conditioned on $B_{S_C}^e$. Then, we may determine the density of the parameters for any complete network structure, simply by performing a change-of-variable operation. For example, consider the complete belief-network structure $x \to y$ for our two-variable domain. A parameter set for the joint space is $\{\theta_{xy}, \theta_{\bar{x}y}, \theta_{x\bar{y}}\}$; and a parameter set for the network structure is $\{\theta_x, \theta_{y|x}, \theta_{y|\bar{x}}\}$. These sets are related by the following relations:

$$\theta_{xy} = \theta_x \theta_{y|x} \quad \theta_{\bar{x}y} = (1-\theta_x)(\theta_{y|\bar{x}}) \quad \theta_{x\bar{y}} = \theta_x(1-\theta_{y|x})$$

Thus, given the density $\rho(\theta_{xy}, \theta_{\bar{x}y}, \theta_{x\bar{y}}|B_{S_C}^e, \xi)$ for the joint space, we may compute the density $\rho(\theta_x, \theta_{y|x}, \theta_{y|\bar{x}}|B_{S_C}^e, \xi)$ using the relation

$$\rho(\theta_x, \theta_{y|x}, \theta_{y|\bar{x}}|B_{S_C}^e, \xi) = J \cdot \rho(\theta_{xy}, \theta_{\bar{x}y}, \theta_{x\bar{y}}|B_{S_C}^e, \xi)$$

where $J$ is the Jacobian of the transformation

$$J = \begin{vmatrix} \partial\theta_{xy}/\partial\theta_x & \partial\theta_{\bar{x}y}/\partial\theta_x & \partial\theta_{x\bar{y}}/\partial\theta_x \\ \partial\theta_{xy}/\partial\theta_{y|x} & \partial\theta_{\bar{x}y}/\partial\theta_{y|x} & \partial\theta_{x\bar{y}}/\partial\theta_{y|x} \\ \partial\theta_{xy}/\partial\theta_{y|\bar{x}} & \partial\theta_{\bar{x}y}/\partial\theta_{y|\bar{x}} & \partial\theta_{x\bar{y}}/\partial\theta_{y|\bar{x}} \end{vmatrix}$$
$$= \theta_x(1-\theta_x) \qquad (11)$$

Given the assumption of parameter modularity, this result extends to any belief-network structure. Namely, in Heckerman et al. (1994, Theorem 2), we show that, given the density of the multinomial parameters for the joint space of $U$ conditioned on $B_{S_C}^e$, we may determine the density of the parameters for *any* network structure. To understand this result, consider the incomplete network structure containing no arc between $x$ and $y$ for our two variable domain. The method for determining the density for the parameters of $B_S$ is illustrated in Figure 2. Given the assumption of parameter independence, we may obtain the densities for $\theta_x$ and $\theta_y$ separately. To obtain the density for $\theta_x$, we identify a complete network structure $B_{SC1}$ such that $x$ has the same parents (namely, none) in both $B_S$ and $B_{SC1}$. Next, using the change-of-variable procedure described in the previous paragraph, we determine the density $\rho(\theta_x|B_{SC1}^e, \xi)$ from $\rho(\theta_{xy}, \theta_{\bar{x}y}, \theta_{x\bar{y}}|B_{S_C}^e, \xi)$. Then, we use the assumption of parameter modularity to obtain $\rho(\theta_x|B_S^e, \xi) = \rho(\theta_x|B_{SC1}^e, \xi)$. In a similar manner, as illustrated in the figure, we obtain the density $\rho(\theta_y|B_S^e, \xi)$.

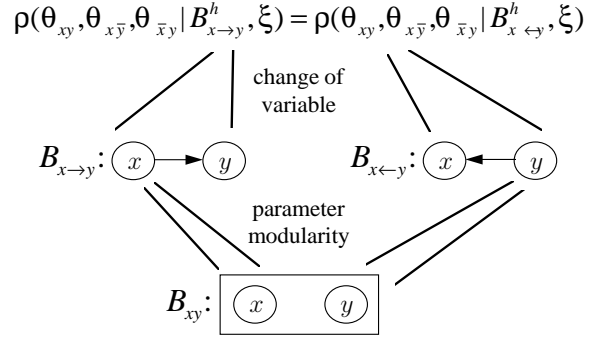

Figure 2: A method for obtaining the density for the parameters of $B_S$ from the density of the joint space of the domain.

Next, let us consider Assumption 5. By a similar argument to that given in the first part of this discussion, we know that given any two complete network structures $B_{SC1}$ and $B_{SC2}$, we may obtain the density $\rho(\Theta_{B_{SC1}}|B_{S_C}^e, \xi)$ from $\rho(\Theta_{B_{SC2}}|B_{S_C}^e, \xi)$, and vice versa, through a change of variable. If we assume that the densities for $B_{SC1}$ are Dirichlet, however, it may not be the case that the densities for $B_{SC2}$ will be Dirichlet. For example, in our two-variable domain, suppose $\rho(\theta_x, \theta_{y|x}, \theta_{y|\bar{x}}|B_{S_C}^e, \xi)$ is equal to a constant (all Dirichlet exponents equal to 1). After a change of variable, from Equation 11 we have

$$\rho(\theta_y, \theta_{x|y}, \theta_{x|\bar{y}}|B_{S_C}^e, \xi) \propto \frac{\theta_y(1-\theta_y)}{\theta_x(1-\theta_x)}$$
$$= \frac{\theta_y(1-\theta_y)}{(\theta_y\theta_{x|y}+(1-\theta_y)\theta_{x|\bar{y}})(1-(\theta_y\theta_{x|y}+(1-\theta_y)\theta_{x|\bar{y}}))}$$

which is not Dirichlet. Consequently, the Dirichlet exponents must be constrained.

In Heckerman et al. (1994, Theorem 7), we show that if $\rho(\Theta_{B_{S_C}}|B_{S_C}^e, \xi)$ is Dirichlet for every complete belief-network structure $B_{S_C}$, then the density of the parameters for the joint space (also conditioned on $B_{S_C}^e$) must also have a Dirichlet distribution. Combining this result with our previous discussion, we see that we may obtain all exponents $N'_{ijk}$ for all belief-network structures, simply by assessing the Dirichlet density for the joint space of $U$ conditioned on $B_{S_C}^e$.

For domains containing a small number of variables, the user may assess this density directly. In larger domains, however, we can use an assessment method based on the notion of an equivalent sample size, described by Winkler (1967). To understand this method, let $\Theta_{x_1,\ldots,x_n}$ denote the set of parameters for the joint space of $U = \{x_1, \ldots, x_n\}$. Denote the Dirichlet density for these parameters as follows:

$$\rho(\Theta_{x_1,\ldots,x_n}|B^e_{S_C},\xi) \propto \prod_{x_1,\ldots,x_n} [\theta_{x_1,\ldots,x_n}]^{(N'_{x_1,\ldots,x_n}-1)} \quad (12)$$

Now, the expectation of $\Theta_{x_1,\ldots,x_n}$ with respect to $\rho(\Theta_{x_1,\ldots,x_n}|B^e_{S_C},\xi)$ is equal to the user's prior probability $p(x_1,\ldots,x_n|B^e_{S_C},\xi)$ for the next instance of the domain to be observed. Thus, using the formula for the expectation of $\Theta_{x_1,\ldots,x_n}$ given the Dirichlet density in Equation 12, we obtain

$$p(x_1,\ldots,x_n|B^e_{S_C},\xi) = \frac{N'_{x_1,\ldots,x_n}}{N'} \quad (13)$$

where

$$N' = \sum_{x_1,\ldots,x_n} N'_{x_1,\ldots,x_n} \quad (14)$$

Thus, we can determine all needed exponents by having a user assess $p(x_1,\ldots,x_n|B^e_{S_C},\xi)$ and $N'$.

The user can assess the joint probability distribution $p(x_1,\ldots,x_n|B^e_{S_C},\xi)$ by constructing a belief network for $U$, given $B^e_{S_C}$. We call this network the user's *prior belief network*.[4]

The constant $N'$ has a simple interpretation as the equivalent number of cases that the user has seen since he was completely ignorant about the domain. Winkler (1967) shows how a user may be trained to assess $N'$.

## 6 The BDe Metric

Given a prior belief network and the constant $N'$, it is not difficult to show that the exponents $N'_{ijk}$ are determined by the relation

$$N'_{ijk} + 1 = N' \cdot p(x_i = k, \Pi_i = j | B^e_{S_C}, \xi) \quad (15)$$

(see Heckerman et al. 1994 for a derivation). This constraint has a simple interpretation in terms of equivalent sample sizes. Namely, $N'_{ij} = \sum_{k=1}^{r_i} N'_{ijk}$ is the equivalent sample size for the parameter set $\Theta_{ij}$—the parameters for $x_i$, given that we have observed the $j$th instance of $\Pi_i$. From Equation 15, we see that

$$N'_{ij} = N' \cdot p(\Pi_i = j | B^e_S, \xi)$$

That is, the equivalent sample size for $\Theta_{ij}$ is just the overall equivalent sample size $N'$ times the probability that we see $\Pi_i = j$.

Substituting Equation 15 into the BD metric (Equation 10), we obtain the BDe metric, a score *e*quivalent metric for belief networks. We note that $N'$ acts as a gain control for learning—the smaller the value of $N'$, the more quickly the BDe metric will favor network structures that differ from the prior belief-network structure.

As an example, let $B_{x \to y}$ and $B_{y \to x}$ denote the belief-network structures where $x$ points to $y$ and $y$ points to $x$, respectively, in our two-variable domain. Suppose that $N' = 12$ and that the user's prior network gives the joint distribution $p(x,y|B^e_{x \to y},\xi) = 1/4, p(x,\bar{y}|B^e_{x \to y},\xi) = 1/4, p(\bar{x},y|B^e_{x \to y},\xi) = 1/6$, and $p(\bar{x},\bar{y}|B^e_{x \to y},\xi) = 1/3$. Using the BDe metric, if we observe database $D$ containing a single case with both $x$ and $y$ true, we obatin

$$p(D, B^e_{x \to y}|\xi) = p(B^e_{x \to y}|\xi) \cdot \frac{11!}{12!} \frac{6!}{5!} \frac{5!}{6!} \frac{3!}{2!}$$

$$p(D, B^e_{y \to x}|\xi) = p(B^e_{y \to x}|\xi) \cdot \frac{11!}{12!} \frac{5!}{4!} \frac{4!}{5!} \frac{3!}{2!}$$

Thus, as required, the BDe metric exhibits the property of score equivalence.[5]

## 7 Causal Networks

People often have knowledge about the causal relationships among variables in addition to knowledge about conditional independence. Such causal knowledge is stronger than is conditional-independence knowledge, because it allows us to derive beliefs about a domain after we intervene. For example, most of us believe that smoking causes lung cancer. From this belief, we infer that if we stop smoking, then we decrease our chances of getting lung cancer. In contrast, if we were to believe that there is only a statistical correlation between smoking and lung cancer, perhaps because there is a gene that causes both our desire to smoke and lung cancer, then we would infer that giving up cigarettes would not decrease our chances of getting lung cancer.

---

[4] At first glance, there seems to be a contradiction in asking the user to construct such a belief network—which may contain assertions of independence—under the assertion that $B^e_{S_C}$ is true. The assertions of independence in the prior network, however, refer to independencies in the next case to be observed. In contrast, the assertion of full dependence $B^e_{S_C}$ refers to long-run fractions.

[5] We note that Buntine presented without derivation the special case of the BDe metric obtained by letting $p(U|B^e_{S_C},\xi)$ be uniform, and noted the property of score equivalence. Also, CH presented a special case of the BD metric wherein each $N'_{ijk}$ is set to 1, yielding a uniform Dirichlet distribution on each density $\rho(\Theta_{ij}|B^e_S,\xi)$. This special case does not exhibit the property of score equivalence.

Causal networks, described—for example—by Spirtes et al. (1993), Pearl and Verma (1991), and Heckerman and Shachter (1994) represent such causal relationships among variables. In particular, a causal network for $U$ is a belief network for $U$, wherein it is asserted that each nonroot node $x$ is caused by its parents. The precise meaning of cause and effect is not important for our discussion. The interested reader should consult the previous references.

More formally, we define a causal network to be a pair $(C_S, C_P)$, where $C_S$ is a causal-network structure and $C_P$ is a set of probability distributions corresponding to that structure. The event $C_S^e$ is the same as that for a belief-network structure, except that we also include in the event the assertion that each nonroot node is caused by its parents.

In contrast to the case of belief networks, it is not appropriate to require the properties of event equivalence or score equivalence. For example, in our two-variable domain, both the causal network $C_{S1}$ where $x$ points to $y$ and the causal network $C_{S2}$ where $y$ points to $x$ represent the assertion that $x$ and $y$ are dependent. The network $C_{S1}$, however, in addition represents the assertion that $x$ causes $y$, whereas the network $C_{S2}$ represents the assertion that $y$ causes $x$. Thus, the events $C_{S1}^e$ are $C_{S2}^e$ are not equal. Indeed, it is reasonable to assume that these events—and the events associated with any two different causal-network structures—are mutually exclusive.

Therefore, the consequences of event equivalence discussed in Section 5 do not apply to causal networks. In particular, the exponents $N'_{ijk}$ have no theoretical constraints, and we may use the BD metric to score causal networks. Nonetheless, for practical reasons, it is useful to constrain the parameters $N'_{ijk}$. SDLC describe one such approach. First, as we do, they asses a prior network. Then, for each variable $x_i$ and each instance $j$ of $\Pi_i$ in the prior network, they allow the user to specify an equivalent sample size $N'_{ij}$. From these assessments, SDLC compute equivalent sample sizes $N'_{ij}$ for other network structures using an *expansion–contraction* procedure. This method has several appealing theoretical properties, but is computationally expensive. CH's specialization of the BD metric, wherein they set each $N'_{ijk}$ to one is efficient, but ignores the prior network. We have explored a simple approach, wherein each $N_{ij}$ is equal to $N''$, a constant. We call this metric the BDu metric ("u" stands for *u*niform equivalent sample sizes). Of course, the BDe metric may also be used to score causal networks.

Note that, in the context of causal networks, the assumption of parameter modularity (Assumption 4) has an appealing justification. Namely, we can imagine that a causal mechanism is responsible for the interaction between each node and its parents. The assumption of parameter modularity then follows from the assumption that the causal mechanisms are independent.

## 8   Limitations of the BDe Metric

Let us again consider the scoring of belief networks. Although our method for determining the exponents $N'_{ijk}$ is simple, it is—in a sense—too simple. Namely, it may be the case that a user has more knowledge about some variables than others, and would like to assess different equivalent sample sizes $N'_{ij}$ for different values of $i$ and $j$. If the network is causal, doing so represents no problem. As we have seen in Section 5, however, doing so in the case of belief networks requires that we abandon at least one of (1) score equivalence, (2) Dirichlet priors, or (3) the ability to score all possible network structures.

We believe it is important to retain score equivalence if at all possible. Furthermore, it is computationally expensive to abandon the Dirichlet assumption. There is promise, however, in avoiding the third assumption. Namely, given a non-score-isomorphic metric that accommodates variable dependent sample sizes, we could use it to score only one element from each equivalence class of isomorphic network structures. To do so, we need a method for designating exactly one network structure from each equivalence class as the network to be scored. A simple approach would be to ask the user to specify a complete ordering over the domain variables. For example, given the ordering $(x, y, z)$ for our three-variable domain, the equivalence class corresponding to the conditional independence of $x$ and $z$ given $y$ would be represented by the network structure $x \to y \to z$; and we would score only this structure. As a more subtle example, given the same ordering, the equivalence class corresponding to the conditional independence of $x$ and $y$ given $z$ would be represented by $x \to z \to y$, because among those network structures in this equivalence class, $x$ occurs first only in this network structure.

## 9   Priors for Network Structures

To complete the information needed to compute our metrics, the user must assess the prior probabilities for the network structures. These assessments are logically independent of the assessment of the prior network, except in the limit as equivalent sample size(s) approach infinity, when the prior network structure must receive a prior probability of one. Nonetheless, structures that closely resemble the prior network tend to have higher prior probabilities.

Here, we propose the following parametric formula for $p(B_S^e|\xi)$ that makes use of the prior network. Given a network structure $B_S$, let $\delta_i$ denote the number of nodes in the symmetric difference of the parents of $x_i$ in $B_S$ and the parents of $x_i$ in the prior network structure. Then, $B_S$ and the prior network differ by $\delta = \sum_{i=1}^n \delta_i$ arcs; and we penalize $B_S$ by a constant factor $0 < \kappa \leq 1$ for each such arc. That is, we set

$$p(B_S^e|\xi) = c \; \kappa^\delta \qquad (16)$$

where $c$ is a normalization constant. This formula is simple, as it requires only the assessment of a single constant $\kappa$. Nonetheless, we can imagine generalizing the formula by punishing different arc differences with different weights, as suggested by Buntine. Although this parametric form does not satisfy score equivalence, we may recover this property, as described in the previous section, by designating within each event equivalence class the network structure to be scored.

## 10 Evaluation

In this section, we evaluate the BDe metric using the 36-node Alarm network for the domain of ICU ventilator management [Beinlich et al., 1989]. In our evaluations we start with the given network, which we call the *gold-standard network*. Next, we generate a database from the given network, using a Monte-Carlo technique. Then, we use one of the scoring metrics and a local search procedure similar to the one described in Lam and Bacchus (1993) to identify a high-scoring network structure. Next, we use the database and prior knowledge to populate the probabilities in the new network, called the *learned network*. In particular, we set each probability $p(x_i = k|\Pi_i = j)$ to be the posterior mean of $\theta_{ijk}$, given the database. Finally, we compare the joint distributions of the gold-standard and learned networks.

In this paper, we use the cross-entropy measure for comparison. In particular, let $q(x_i, \ldots, x_n)$ and $p(x_i, \ldots, x_n)$ denote the probability of an instance of $U$ obtained from the gold-standard and learned networks, respectively. Then we measure the accuracy of a learning algorithm using the cross entropy $H(q, p)$, given by

$$H(q, p) = \sum_{x_1, \ldots, x_n} q(x_i, \ldots, x_n) \; \log \frac{q(x_i, \ldots, x_n)}{p(x_i, \ldots, x_n)} \qquad (17)$$

The lower the value of the cross entropy, the more accurate the algorithm. In Heckerman et al. (1994), we describe a method for computing the cross entropy of two networks that makes use of the network structures.

In our experiments, we construct prior networks by adding noise to the gold-standard network. We control the amount of noise with a parameter $\eta$. When $\eta = 0$, the prior network is identical to the gold-standard network, and as $\eta$ increases, the prior network diverges from the gold-standard network. When $\eta$ is large enough, the prior network and gold-standard networks are unrelated. To generate the prior network, we first add $2\eta$ arcs to the gold-standard network, creating network structure $B_{S1}$. When we add an arc, we copy the probabilities in $B_{P1}$ so as to maintain the same joint probability distribution for $U$. Next, we perturb each conditional probability in $B_{P1}$ with noise. In particular, we convert each probability to log odds, add to it a sample from a normal distribution with mean zero and standard deviation $\eta$, convert the result back to a probability, and renormalize the probabilities. Then, we create another network structure $B_{S2}$ by deleting $\eta$ arcs and reversing up to $2\eta$ arcs (a reversal may create a directed cycle, in which case, the reversal is not done). Next, we perform inference using the joint distribution determined by network $(B_{S1}, B_{P1})$ to populate the conditional probabilities for network $(B_{S2}, B_{P2})$, which we return as the prior network.

Figure 3 shows the cross entropy of learned networks with respect to the Alarm network (inverse learning accuracy) as a function of the deviation of the prior-network from the gold- standard network ($\eta$) and the user's equivalent sample size ($N'$) for the BDe metric. In this experiment, we used 100-case databases generated from the Alarm network. For each value of $\eta$ and $N'$, the cross-entropy values shown in the figure represent an average over ten learning instances, where in each instance we used a different database and prior network. The databases and prior networks generated for a given value of $\eta$ were used for all values of $N'$. We made the prior parameter $\kappa$ a function of $N'$—namely, $\kappa = 1/(N' + 1)$—so that it would take on reasonable values at the extremes of $N'$. (When $N' = 0$, reflecting complete ignorance, all network structures receive the same prior probability. Whereas, in the limit as $N'$ approaches infinity, reflecting complete confidence, the prior network structure receives a prior probability of one.)

The qualitative behavior of the curve is reasonable. Namely, when $\eta = 0$—that is, when the prior network was identical to the Alarm network—learning accuracy increased as the equivalent sample size $N'$ increased. Also, learning accuracy decreased as the prior network deviated further from the gold-standard network, demonstrating the expected result that prior knowledge is useful. In addition, when $\eta \neq 0$, there was a value of $N'$ associated with optimal accuracy. This result is not surprising. If $N'$ is too large, then the deviation between the true values of the parameters and their priors degrade performance. On the other hand, if $N'$ is too small, the metric is ignoring useful prior knowledge. We speculate that results of this kind can be used to calibrate users in the assessment of $N'$.

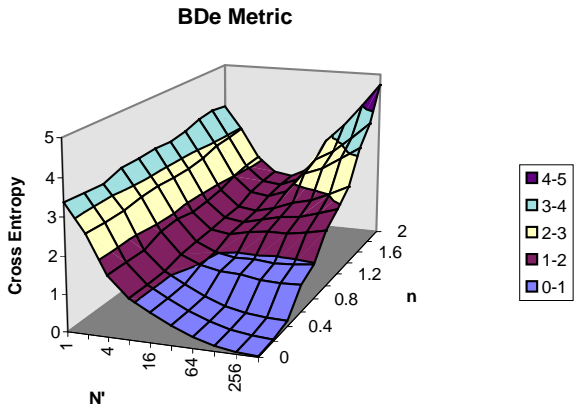

Figure 3: Evaluation results.

Also, the quantitative results are encouraging. To provide a scale for cross entropy in the Alarm domain, note that the cross entropy of the Alarm network with an empty network for the domain (i.e., a network where all variables are independent) whose marginal probabilities are determined from the Alarm network is 13.6. Using only a 100 case database, and a prior network with a significant amount of noise—$\eta = 2$, the cross entropy for the BDe metric, at the optimum value of $N'$ ($= 16$), is only 1.6.

## Acknowledgments

We thank Jack Breese, Wray Buntine, Greg Cooper, Steffen Lauritzen, and anonymous reviewers for useful suggestions.